\documentclass[lettersize,journal]{IEEEtran}
\usepackage{amsmath,amsfonts}
\usepackage{array}
\usepackage[caption=false,font=footnotesize,labelfont=rm,textfont=rm]{subfig}
\usepackage{textcomp}
\usepackage{stfloats}
\usepackage{url}
\usepackage{verbatim}
\usepackage{graphicx}
\usepackage{cite}
\usepackage{booktabs}
\usepackage{multirow}
\usepackage{color}
\usepackage{pifont}
\usepackage{makecell} 
\usepackage{marvosym} % \MVRIGHTarrow

\usepackage[linesnumbered,ruled,vlined]{algorithm2e}
\usepackage{algpseudocode}  
\usepackage{colortbl}
\usepackage{pifont}
\usepackage[dvipsnames, svgnames, x11names]{xcolor}% 颜色支持
\usepackage[colorlinks=true,linkcolor=red,citecolor=green,urlcolor=magenta]{hyperref}
\definecolor{LightCyan}{rgb}{0.88,1,1}
\begin{document}

\title{Rethinking Representations for Cross-Domain Infrared Small Target Detection: A Generalizable Perspective from the Frequency Domain} 

\author{Yimin Fu, Songbo Wang, Feiyan Wu, Jialin Lyu, Zhunga Liu, Michael K. Ng
\thanks{

Yimin Fu and Michael K. Ng are with the Department of Mathematics, Hong Kong Baptist University, Hong Kong, China (e-mail: fuyimin@hkbu.edu.hk; michael-ng@hkbu.edu.hk).

Songbo Wang, Jialin Lyu, and Zhunga Liu are with the School of Automation, Northwestern Polytechnical University, Xi'an, 710072, China (e-mail: wangsongbo@mail.nwpu.edu.cn; jialinlv@mail.nwpu.edu.cn; liuzhunga@nwpu.edu.cn).

Feiyan Wu is with the Wangxuan Institute of Computer Technology, Peking University, Beijing, China (wufeiyan@pku.edu.cn).
}}

\markboth{Manuscript Under Review}%
{Shell \MakeLowercase{\textit{et al.}}: A Sample Article Using IEEEtran.cls for IEEE Journals}
\maketitle

\begin{abstract}
The accurate target–background separation in infrared small target detection (IRSTD) highly depends on the discriminability of extracted representations.
However, most existing methods are confined to domain-consistent settings, while overlooking whether such discriminability can generalize to unseen domains.
In practice, distribution shifts between training and testing data are inevitable due to variations in observational conditions and environmental factors.
Meanwhile, the intrinsic indistinctiveness of infrared small targets aggravates overfitting to domain-specific patterns.
Consequently, the detection performance of models trained on source domains can be severely degraded when deployed in unseen domains.
To address this challenge, we propose a spatial–spectral collaborative perception network (S$^2$CPNet) for cross-domain IRSTD.
Moving beyond conventional spatial learning pipelines, we rethink IRSTD representations from a frequency perspective and reveal inconsistencies in spectral phase as the primary manifestation of domain discrepancies. 
Based on this insight, we develop a phase rectification module (PRM) to derive generalizable target awareness.
Then, we employ an orthogonal attention mechanism (OAM) in skip connections to preserve positional information while refining informative representations.
Moreover, the bias toward domain-specific patterns is further mitigated through selective style recomposition (SSR).
Extensive experiments have been conducted on three IRSTD datasets, and the proposed method consistently achieves state-of-the-art performance under diverse cross-domain settings.
The code will be released at \url{https://github.com/fuyimin96/S2CPNet} upon acceptance.
\end{abstract}  


\section{Introduction}
\IEEEPARstart{I}{nfrared} small target detection (IRSTD)~\cite{zhao2022single} plays a crucial role in various military and civilian applications.
Although infrared sensing~\cite{morris2007statistics,he2021infrared,zhang2021deep} offers robust perception under adverse environments, its unique characteristics also pose additional challenges to IRSTD compared with generic object detection~\cite{liu2020deep,zou2023object,liu2024crada}.
On the one hand, targets captured at long distances appear extremely small and structurally diverse. On the other hand, infrared targets exhibit intrinsic visual indistinctiveness and low signal-to-clutter ratios (SCRs). These challenges jointly complicate the extraction of discriminative representations, leading to undesired missed detections and false alarms.

\begin{figure}[]
\centering 
\includegraphics[width=0.5\textwidth]{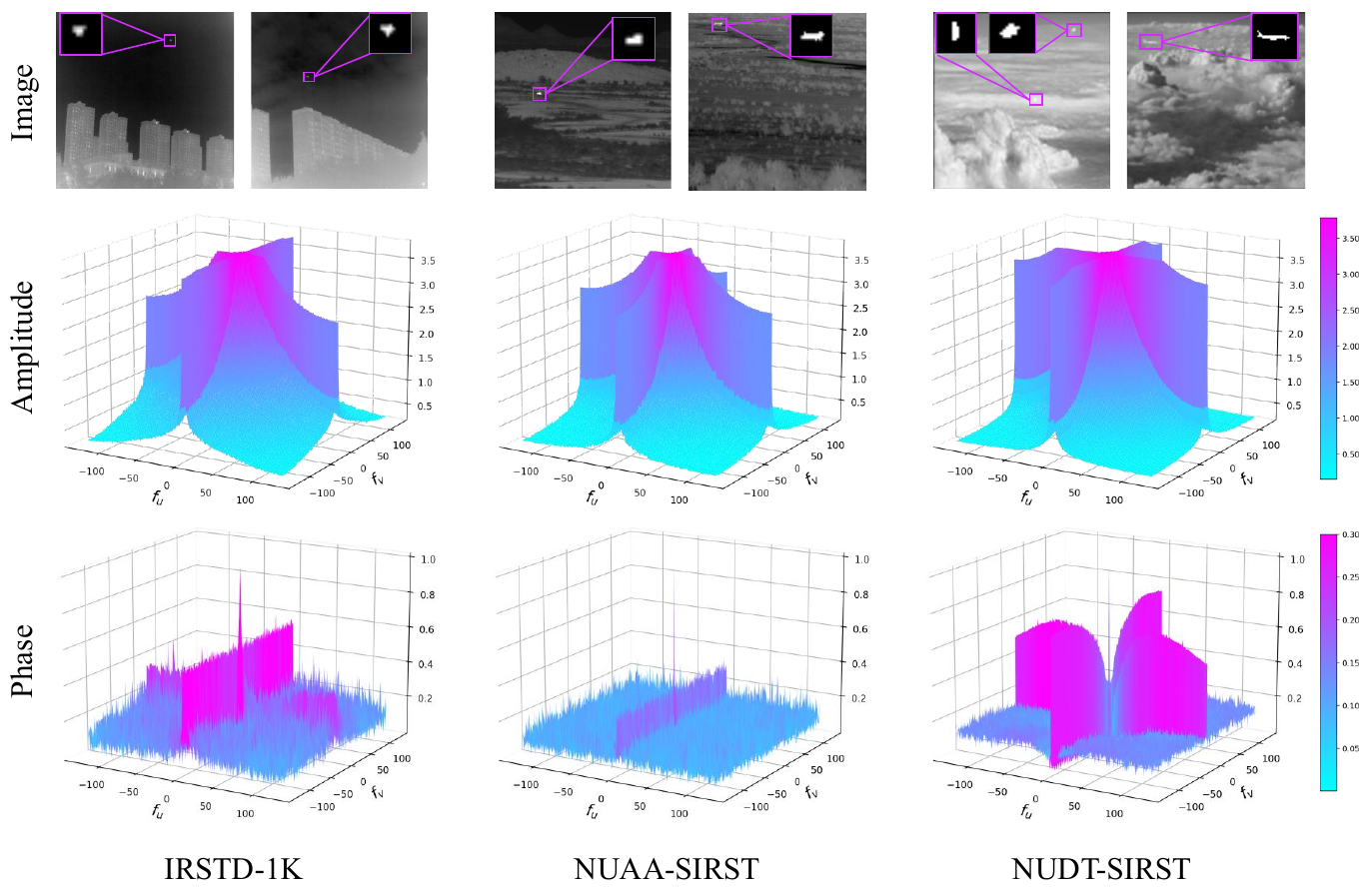}
\caption{Representative image examples and overall frequency characteristic of different IRSTD datasets.}
\label{fre_ana} 
\end{figure} 

Early IRSTD research primarily follows a model-driven paradigm, which discovers targets from backgrounds using various image processing techniques~\cite{tophat, MPCM, MSLSTIPT}.
Although computationally efficient and interpretable, the heavy reliance on handcrafted features and parameter tuning severely limits their applicability in dynamic and complex scenarios.
Benefiting from the powerful feature extraction ability of deep neural networks (DNNs)~\cite{he2016deep,vit,fu2025reason}, data-driven methods have become the mainstream.
Driven by this trend, various model architectures~\cite{DNANet, UIUNet, AGPCNet, ISNet} and learning strategies~\cite{ACMNet, ABC, L2SKNet} have been designed to improve representation discriminability for more precise target–background separation.
However, most existing methods adhere to an idealized single-domain setting, where training and testing data are assumed to follow an identical distribution.
In real-world applications, variations in observational conditions and environmental factors inevitably induce distribution shifts~\cite{zhang2021adaptive,wang2022generalizing}.
Such shifts often cause severe performance degradation, thereby imposing imperative requirements on the cross-domain generalization of learned representations~\cite{oza2023unsupervised,schwonberg2025domain}.

Although domain adaptation (DA) techniques have been widely integrated into detectors to transfer knowledge across domains~\cite{oza2023unsupervised}, the reliance on pre-collected unlabeled data in target domains severely constrains their applicability in open and dynamic environments.
In contrast, domain generalization (DG)~\cite{wang2022generalizing} offers a more practical solution, which acquires underlying task understanding without requiring target-domain data.
Unfortunately, the substantial differences between infrared and natural images undercut the prospect of directly applying DG approaches in the general computer vision field to IRSTD.
Moreover, the intrinsic spatial indistinctiveness of infrared small targets~\cite{yang2025deep} further complicates the extraction of discriminative representations, rendering models prone to being trapped in domain-specific shortcut patterns.
This necessitates rethinking representations in cross-domain IRSTD and seeking new pathways beyond conventional spatial learning pipelines to reconcile discriminability with generalization.
Since frequency components can unveil spatially non-intuitive cues for model generalization~\cite{wang2020high}, we turn to the frequency domain to dissect the essence of domain discrepancies in IRSTD.
According to the empirical Fourier analysis of different IRSTD datasets (Fig.~\ref{fre_ana}), the magnitude spectra remain largely consistent across domains, whereas the phase congruency exhibits significant inter-domain variations.
This observation distinctly reveals that domain discrepancies in IRSTD arise mainly from inconsistencies in spectral phase.
Therefore, promoting phase alignment across domains offers a principled way to enhance the generalization of IRSTD representations.

Building upon this insight, we propose a spatial–spectral collaborative perception network (S$^2$CPNet) for cross-domain IRSTD.
Specifically, a phase recitification module~(PRM) is jointly operated with spatial convolutional blocks throughout the feature encoding stage.
Introducing phase congruency variation simulation implicitly regularizes the model to rectify its over-dependence on domain-specific patterns, thereby facilitating the derivation of generalizable target awareness.
Then, during decoding, cross-stage representations are progressively refined via an orthogonal attention mechanism (OAM), which simultaneously alleviates the positional degradation in skip connections.
Moreover, the style characteristics across different domains are dynamically abstracted into prototypes, upon which the bias toward domain-specific patterns is further mitigated through selective style recomposition (SSR).
Finally, we conduct cross-domain evaluations on three IRSTD datasets, and S$^2$CPNet consistently achieves state-of-the-art performance under diverse settings.

Our main contributions can be summarized as follows:
\begin{itemize}
\item We propose S$^2$CPNet, a cross-domain generalized IRSTD framework that rethinks IRSTD representations from the frequency perspective.
During encoding, biases toward domain-specific patterns are effectively alleviated through the incorporation of a phase rectification module.

\item We design an orthogonal attention mechanism to facilitate the preservation of positional information.
This effectively mitigates the spatial ambiguity problem when refining cross-stage representations during decoding.

\item We develop a selective style recomposition strategy, where the application of identity-independent projection is adaptively guided by domain style prototypes, thereby further reducing distribution gaps across domains.
\end{itemize}

The rest of this paper is structured as follows. Section~\ref{sec2} reviews prior studies related to our work. Section~\ref{sec3} presents the problem formulation and technical details of the proposed method. Section~\ref{sec4} reports comprehensive experimental evaluations and the corresponding analyses. 
Finally, Section~\ref{sec5} concludes this paper and outlines future research directions.

\section{Related Works}\label{sec2}
\subsection{Infrared Small Target Detection}
Depending on the input configuration, IRSTD can be categorized into single-frame~\cite{zhao2022single} and multi-frame~\cite{chen2024sstnet,wu2025neural} schemes.
Although multi-frame integration provides richer context, single-frame IRSTD remains predominant due to computational efficiency and real-time applicability.

Early model-driven IRSTD methods employ various image processing techniques for target–background separation, including filtering~\cite{tophat, LIG}, local contrast enhancement~\cite{MPCM, RLCM, TLLCM}, and tensor decomposition~\cite{MSLSTIPT,zhang2019infrared}.
However, their reliance on hand-crafted features and manual parameter tuning leads to unsatisfactory performance in complex scenarios.
With the powerful representation capacity of DNNs and the growing accessibility of training resources, current research in IRSTD mainly follows the data-driven paradigm.
Concretely, various learning strategies have been developed to extract discriminative target representations.
Dai et al.~\cite{ACMNet} facilitated scene understanding through asymmetric contextual integration.
Zhang et al.~\cite{ISNet} enhanced the discriminability of target representations by reconstructing edge and shape structures.
Following these efforts, a series of studies subsequently emerged to further enhance the capture of critical patterns.
Yuan et al.~\cite{SCTransNet} aggregated contextual information across multiple levels to improve the discrimination between targets and backgrounds.
Wu et al.~\cite{L2SKNet} designed a local saliency kernel module to strengthen the extraction of discriminative target features.
By enhancing sensitivity to locations and scales, Liu et al.~\cite{liu2024infrared} proposed a multi-scale detection head to achieve more comprehensive target perception.
Lin et al.~\cite{lin2023learning} integrated shape priors into the training process to enable accurate target detection against complex backgrounds.
In addition, different customized model architectures have also been designed to prevent small targets from being submerged in background clutter, such as dense nested attention~\cite{DNANet}, hierarchical U-shape structure~\cite{UIUNet}, pyramid contextual integration~\cite{AGPCNet}, atrous residual
block~\cite{MFFSANet26}, and hybrid feature encoder~\cite{MTUNet}.
Following the rise of foundation models, knowledge from pretrained models has also been leveraged to boost IRSTD performance~\cite{zhang2024irsam,zhang2025saist,fu2025unified}.
To advance the deployment in real-world applications, several studies have explored IRSTD under weakly supervised~\cite{ying2023mapping,yu2025easyhard} and noise-interference~\cite{yuan2024irstdid} settings.
Although some recent studies~\cite{chi2025contrast,Liu2026HSDW} have attempted to improve the generalization of IRSTD, their exploration is limited to single-domain scenarios.
In addition, their method designs still follow conventional spatial learning pipelines, which leaves substantial room for further performance improvement.

\subsection{Cross-Domain Segmentation}
Unlike bounding-box prediction in generic object detection, IRSTD formulates outputs as segmentation masks to capture both target location and morphology.
Although domain adaptation (DA)~\cite{kouw2019review} has been widely adopted for cross-domain segmentation, the dependence on unlabeled target data severely constrains its deployment in real-world applications.
In contrast, domain generalization (DG)~\cite{wang2022generalizing} provides a more practical solution that acquires universal task understanding without requiring target-domain data.

Existing domain generalized segmentation approaches generally follow two lines: input augmentation~\cite{yue2019domain,peng2021global,li2022learning,jiang2023domain}and feature manipulation~\cite{pan2018two,pan2019switchable,bi2024learning}.
Input augmentation–based approaches aim to synthesize images with diverse distributions, thereby encouraging the model to learn domain-invariant representations.
The typical used augmentation operations include style manipulation~\cite{yue2019domain}, texture randomization~\cite{peng2021global}, and image transformations~\cite{jiang2023domain}.
Different from the former, feature manipulation-based approaches alleviate overfitting to domain-specific information by imposing constraints on feature representations.
Li et al.~\cite{li2024faa} designed a hybrid feature augmentation and alignment strategy to achieve robust object understanding across diverse scenarios.
Pan et al.~\cite{pan2018two,pan2019switchable} successively introduced instance normalization and instance whitening to regularize DNNs against appearance variations in content-related representation learning. 
Following these, subsequent improvements have been made in terms of feature discrimination~\cite{choi2021robustnet}, semantic awareness~\cite{peng2022semantic}, and representation consistency~\cite{zhao2022style}.
More explicitly, Bi et al.~\cite{bi2024learning} designed a mask attention mechanism to select domain-independent contents.
In addition, multiple variants of contrastive loss~\cite{lee2022wildnet,huang2023style} have been employed for robust segmentation in unseen target domains.

\begin{figure*}[]
\centering 
\includegraphics[width=\textwidth]{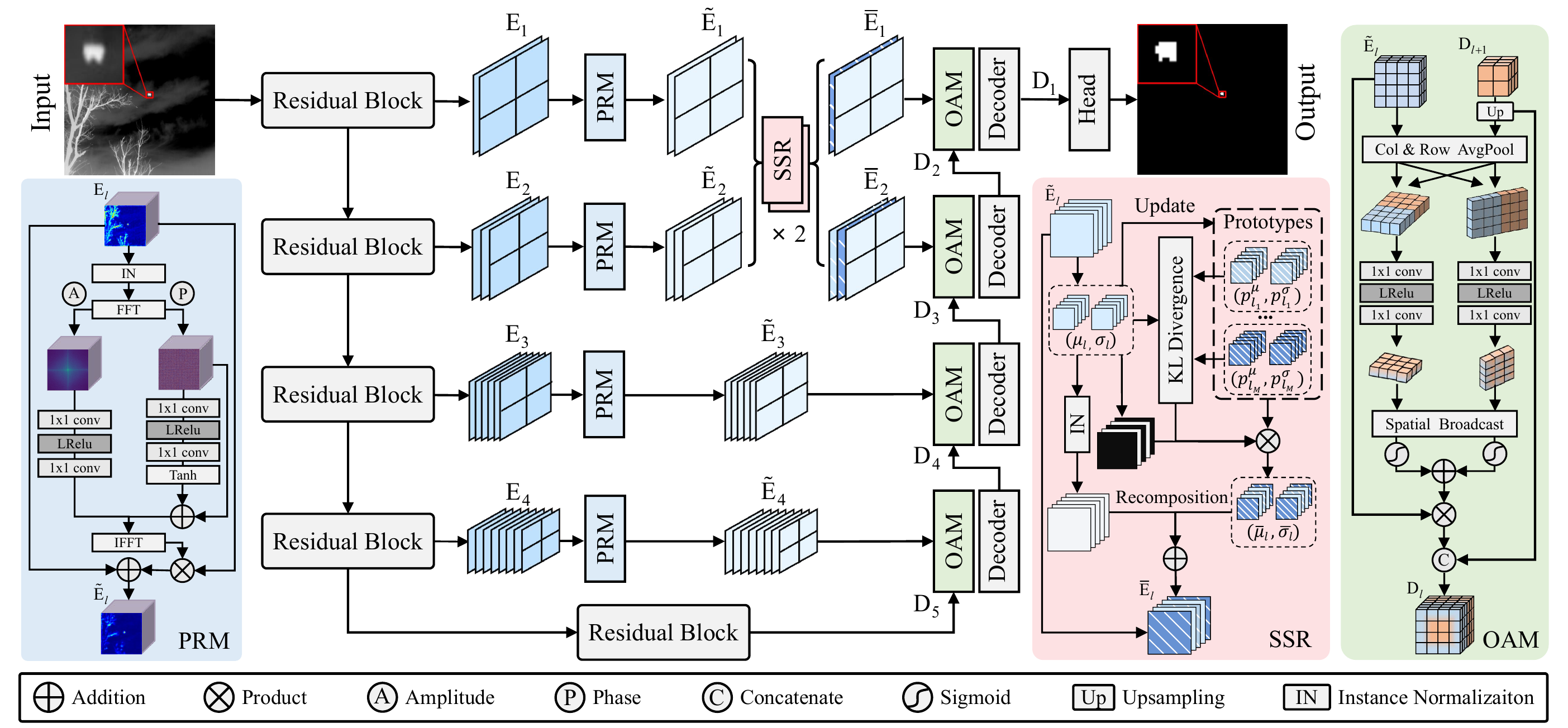}
\caption{The architecture overview of our proposed S$^2$CPNet. Three key components are employed in the encoder-decoder framework: phase rectification module, orthogonal attention mechanism, and selective style recomposition.}
\label{s2cp_net}
\end{figure*}

\section{Methodology}\label{sec3}
In this section, we first introduce the problem formulation of cross-domain IRSTD. Then, an overview of our proposed S$^2$CPNet is provided, followed by technical details of each main component.

\subsection{Problem Formulation and Overview}
Given training data from source domains $S=\{(x^s, y^s)\}$, where the location of targets in each infrared image $x^s\in\mathbb{R}^{ H \times W \times C}$ with spatial resolution $H \times W$ is indicated by the corresponding ground-truth mask $y^s\in \{0,1\}^{H \times W}$.
The objective of the cross-domain IRSTD is to train a detector $\phi(\cdot)$ on source-domain data that generalizes well on unseen target domains $\mathcal{T}=\{(x^t, y^t)\}$, which can be formulated as minimizing the following risk:
\begin{equation}
R
= \underset{\phi}{\operatorname{min}}\;
\mathbb{E}_{(x^t,y^t)\sim\mathcal{T}}
\Big[\sum_{h=1}^{H}\sum_{w=1}^{W}
y^t\log(\phi(x^t))
\Big],
\end{equation}
where $\phi(x^t)$ denotes the prediction at each location.

The overall architecture of the proposed S$^2$CPNet is illustrated in Fig.~\ref{s2cp_net}.
Built upon an $L$-stage encoder–decoder framework, multi-level feature maps $E=[E_1, \ldots, E_L]$ are extracted through cascaded convolutional blocks coupled with the phase rectification module.
Then, the spatial details are gradually reconstructed from cross-stage representations refined by the orthogonal attention mechanism.
Based on the identity-independent projection, selective style recomposition is conducted during the early encoding stages for further generalization improvements.

\subsection{Phase Rectification Module}
The discriminability of representations learned from source domains cannot be maintained on unseen domains due to distribution shifts, and the intrinsic spatial indistinctiveness of infrared small targets further exacerbates the overfitting risk of spatial learning pipelines to domain-specific patterns.
Since domain discrepancies are predominantly manifested through disparities in spectral phase, unifying phase congruency in the frequency domain provides a principled way to improve generalization.

To this end, we integrate phase recitification modules with convolutional blocks during the encoding stage. 
Specifically, at the $l^{th}$ stage, feature maps $E_l$ extracted from the convolutional block are transformed into amplitude and phase components in the frequency domain by a 2-D Fast Fourier Transform (FFT):
\begin{equation}
\{A_l, P_l\} = \text{FFT}(\text{BN}(E_l)),
\end{equation}
where $\text{BN}(\cdot)$ denotes batch normalization.
Then, the diversities in phase congruency are simulated through a modulation branch and imposed on the original phase component:
\begin{equation}
\tilde{P}_l = {P}_l+\tanh\big(W_{l_2}^{P}(\operatorname{LReLU}(W_{l_1}^{P}(P_l)))\big),
\end{equation}
where $\operatorname{LReLU}(\cdot)$ denote the Leaky ReLU activation, and $W_{l_1}^{P}, W_{l_2}^{P}$ are $1 \times 1$ convolutional operations.
This encourages the model to rectify phase deviations under varying observational conditions, thereby enhancing the robustness against domain shifts.
In parallel, the target saliency encoded in the amplitude components is extracted according to the energy cues:
\begin{equation}
\tilde{A}_l = W_{l_2}^{A}(\operatorname{LReLU}(W_{l_1}^{A}( A_l))),
\end{equation}
where $W_{l_1}^{A}$ and $W_{l_2}^{A}$ are $1 \times 1$ convolutional operations.
Subsequently, a perception indication map $M_l$ can be obtained by Inverse Fast Fourier Transform (IFFT):
\begin{equation}
M_l=\text{IFFT}(\tilde{A}_l,\tilde{P}_l),
\end{equation}
which is used to guide the extraction of representations for deriving generalizable target awareness:
\begin{equation}
\tilde{E}_{l}=\left(M_l \odot E_l+E_l\right).
\end{equation}
where $\odot$ represents the element-wise multiplication.

\subsection{Orthogonal Attention Mechanism}
After rectifying the over-dependence on domain-specific patterns, the accurate target perception therefore hinges on effective resolution recovery, which is typically realized through skip connections.
While various attention mechanisms~\cite{hu2018squeeze,wang2022uctransnet} have been employed to refine cross-stage representations, this process also introduces spatial ambiguity due to the loss of positional information.

To overcome this limitation, we design an orthogonal attention mechanism to mitigate positional degradation during resolution recovery.
In the $l^{th}$ decoding stage, the output feature maps $D_{l+1}$ from the previous decoding block are first upsampled to ${D}_{l}^{\prime}$ to match the spatial dimension of the corresponding encoded feature maps $\tilde{E}_{l} \in \mathbb{R}^{ H_l \times W_l \times C_l}$. 
Then, unlike conventional attention mechanisms that use 2-D global pooling to model channel-wise importance, we perform spatial squeeze in an orthogonal manner:
\begin{equation}
\begin{aligned}
\mathcal{G}_{l_H}^E=\frac{1}{W_l} \sum_{w=1}^{W_l} \tilde{E}_l[:, w,:], \quad \mathcal{G}_{l_W}^E=\frac{1}{H_l} \sum_{h=1}^{H_l}  \tilde{E}_l[h,:,:],\\
\mathcal{G}_{l_H}^D=\frac{1}{W_l} \sum_{w=1}^{W_l} {D}_{l}^{\prime}[:, w,:], \quad \mathcal{G}_{l_W}^D=\frac{1}{H_l}  \sum_{h=1}^{H_l}  {D}_{l}^{\prime}[h,:,:],
\end{aligned}
\end{equation}
and $\mathcal{G}_{l_H}^E, \mathcal{G}_{l_H}^D \in \mathbb{R}^{ H_l \times 1 \times C_l}$ and $\mathcal{G}_{l_W}^E, \mathcal{G}_{l_W}^D \in \mathbb{R}^{ 1 \times W_l \times C_l}$ are subsequently concatenated into importance descriptors in horizontal and vertical directions:
\begin{equation}
\mathcal{G}_{l_H}=[\mathcal{G}_{l_H}^E,\mathcal{G}_{l_H}^D], \quad \mathcal{G}_{l_W}=[\mathcal{G}_{l_W}^E,\mathcal{G}_{l_W}^D],
\end{equation}
where $[\cdot,\cdot]$ denotes the concatenation along the channel dimension.
Thereafter, the spatial dependencies along each direction are concurrently modeled and integrated into the attention mask:
\begin{equation}
\mathcal{G}_{l}=\sum_{d \in \{H,W\}} W_{l_2}^{d}(\operatorname{LReLU}(W_{l_1}^{d}(\mathcal{G}_{l_d}))),
\end{equation}
where $W_{l_1}^{d}$ and $W_{l_2}^{d}$ are $1 \times 1$ convolutional operations.
Finally, the output features can be generated from the refined cross-stage representations:
\begin{equation}
D_{l}=\mathcal{F}_{l}([\zeta(\mathcal{G}_{l}) \odot \tilde{E}_{l},{D}_{l}^{\prime}]),
\end{equation}
where $\zeta(\cdot)$ and $\mathcal{F}_{l}(\cdot)$ represent the sigmoid function and decoding block.

\subsection{Selective Style Recomposition}
In domain generalized segmentation, a common strategy to reduce domain gaps is to impose statistical regularization~\cite{pan2018two,pan2019switchable} on feature distributions. 
However, uniformly enforcing such regularizations across entire feature maps may inadvertently suppress certain discriminative cues entangled with domain-specific style characteristics.

To mitigate this impairment, we apply selective style recomposition during the early encoding stage.
Specifically, the channel-wise feature statistics (i.e., mean and standard deviation) of the rectified feature maps $\tilde{E}_{l}\in \mathbb{R}^{ H_l \times W_l \times C_l}$ are calculated as $\mu_l \in \mathbb{R}^{C_l}$ and $\sigma_l \in \mathbb{R}^{C_l}$.
Then, following~\cite{huang2023style}, the style characteristics of $M$ source domains are abstracted into prototypes $\mathcal{P}_l = \{(p_{l_m}^{\mu}, p_{l_m}^{\sigma})\}_{m=1}^{M}$, among which corresponding to the domain identity of $E_l$ (denoted as ${m^*}$) is dynamically updated during training:
\begin{equation}
\begin{aligned}
& p_{l_{m^*}}^\mu=\alpha p_{l_{m^*}}^\mu+(1-\alpha) \mu_l, \\
& p_{l_{m^*}}^\sigma=\alpha p_{l_{m^*}}^\sigma+(1-\alpha) \sigma_l,
\end{aligned}
\end{equation}
where the momentum factor $\alpha$ is set to 0.95.
Afterward, the style attribution coefficients ${\kappa}=\{\kappa_m\}_{m=1}^M$ can be estimated according to the KL-divergence between the statistical distribution of feature maps and each domain prototype:
\begin{equation}
\kappa_{l_m} =
\frac{
\exp\!\big(-\,D_{KL}(\mathcal{N}(\mu_l,\sigma_l^2) \| \mathcal{N}(p_{l_m}^{\mu},{p_{l_m}^{\sigma^2}} ))\big)
}{
\sum\nolimits_{M}
\exp\!\big(-\,D_{KL}(\mathcal{N}(\mu_l,\sigma_l^2) \| \mathcal{N}(p_{l_m}^{\mu},{p_{l_m}^{\sigma^2}} ))\big)
}.
\end{equation}
Next, we employ the variance as the indicator of sensitivity to style variations, based on which an activation mask $\mathcal{I}_l=\{I_{l_c}\}_{c=1}^C$ is determined by:
\begin{equation}
I_{l_c}=1\{c \in \operatorname{argsort}_{{top- }\tau}(\mu_l)\}, %c\in\{1,\ldots,C\}
\end{equation}
where $\tau\in(0,1]$ is used to control the proportion of channels on which an identity-independent projection is selectively applied:
\begin{equation}
(\bar{\mu}_l(c),\bar{\sigma}_l(c)) =
\begin{cases}
\displaystyle \sum_{m=1}^{M} (\kappa_{l_m}\, p_{m}^{\mu}(c), \kappa_{l_m}\, p_{m}^{\sigma}(c)), & \text{if } I_{l_c}=1, \\
(\mu_l(c),\sigma_l(c)), & \text{otherwise}.
\end{cases}
\end{equation}
Finally, the style recomposed representations can be formulated in a residual manner:
\begin{equation}\label{style_recomp}
\bar{E}_l = \lambda\tilde{E}_{l}+(1-\lambda)(\bar\sigma_l \frac{\tilde{E}_{l}-\mu_l}{\sigma_l}+\bar\mu_l),
\end{equation}  
where $\lambda$ is a hyperparameter to control the preservation degree of initial representations.

\section{Experiments}\label{sec4}
\subsection{Experimental Setup}
\subsubsection{Dataset}
We evaluate the effectiveness of the proposed method on three public datasets: NUAA-SIRST~\cite{ACMNet}, NUDT-SIRST~\cite{DNANet}, and IRSTD-1K~\cite{ISNet}, which contain 427, 1327, and 1000 images, respectively.
These datasets include various types of infrared small targets captured from diverse background scenes.
Specifically, two cross-domain IRSTD scenarios are investigated: 
\begin{itemize}
\item \textbf{Single source domain}:
Models are trained on a single source dataset and evaluated on the others.
\item \textbf{Multiple source domains}:
Each dataset is alternately used as the target domain while the other datasets serve as source domains.
\end{itemize}

\begin{table*}[t]
\footnotesize 
\renewcommand\arraystretch{1.1}
\setlength{\tabcolsep}{1pt}
\centering
\caption{Additional evaluation of IRSTD performance in the single-domain generalization scenario in terms of IoU (\%), $P_{d}$ (\%), and $F_{a}$ ($10^{-6}$). The best results are highlighted in bold.}
\begin{tabular}{l|c| c c c| c c c| c c c| c c c| c c c| c c c}
\toprule
\multirow{2}{*}{Method}  &\multirow{2}{*}{Venue}&
\multicolumn{3}{c|}{NUDT \MVRightarrow{} IRSTD-1K} &
\multicolumn{3}{c|}{NUDT \MVRightarrow{} NUAA} &
\multicolumn{3}{c|}{NUAA \MVRightarrow{} IRSTD-1K} &
\multicolumn{3}{c|}{NUAA \MVRightarrow{} NUDT} &
\multicolumn{3}{c|}{IRSTD-1K \MVRightarrow{} NUDT} &
\multicolumn{3}{c}{IRSTD-1K \MVRightarrow{} NUAA }\\
\cline{3-20} & & IoU $\uparrow$  & $P_d\uparrow$ & $F_a\downarrow$  & IoU $\uparrow$  & $P_d\uparrow$ & $F_a\downarrow$  & IoU $\uparrow$  & $P_d\uparrow$ & $F_a\downarrow$ & IoU $\uparrow$  & $P_d\uparrow$ & $F_a\downarrow$ & IoU $\uparrow$  & $P_d\uparrow$ & $F_a\downarrow$ & IoU $\uparrow$  & $P_d\uparrow$ & $F_a\downarrow$  \\
\midrule
Tophat~\cite{tophat}    & $\text{OE}_{96}$ & 5.95 & 69.69 &  621.38 & 27.80  & 90.74  & 193.67 & 5.95  & 69.69  & 621.39  & 12.59  & 69.41  & 76.50  & 12.59  & 69.41  & 76.50  & 27.80  & 90.74  & 193.67   \\
MPCM~\cite{MPCM}    & $\text{PR}_{16}$  &  22.90 & 79.12 &  135.53 & 45.18 & 90.74 & 26.25 &  22.90  & 79.12  & 135.53  & 15.60  & 58.62  & 185.54  & 15.60  & 58.62  & 185.54  & 45.18  & 90.74  & 26.25  \\
RLCM~\cite{RLCM}     & $\text{GRSL}_{18}$ & 17.67 & 64.64 &  54.05& 27.16 & 84.25 & 135.79 &  17.67  & 64.64  & 54.05  & 15.38  & 76.40  & 480.03  & 15.38  & 76.40  & 480.03  & 27.16  & 84.25  & 135.79    \\
LIG~\cite{LIG}  & $\text{IPT}_{18}$& 11.67 & 81.14 &  624.59 & 30.10 & 88.88 & 77.39 &  11.67  & 81.14  & 624.59  & 28.29  & 78.41 & 170.65  & 28.29  & 78.41 & 170.65  & 30.10  & 88.88  & 77.39   \\
ACMNet~\cite{ACMNet}   &$\text{WACV}_{21}$ & 40.39 & 83.71 & 104.18 & 59.81 & 92.12 & 20.40 & 37.48 & 74.56 & 158.04 & 36.25 & 74.23 & 138.43 & 36.67 & 73.42 & 66.71 & 64.55 & 90.80 & 19.40     \\
DNANet~\cite{DNANet}     & $\text{TIP}_{22}$ & 46.95 & 80.28 & 57.83 &61.82 & 91.74 & 15.25&  53.58 & 79.47 & 60.54 & 45.55 & 71.55 & 48.14 & 43.11 & 72.73 & 62.43 & 70.80 & 95.49 & 14.54   \\
UIUNet~\cite{UIUNet}       & $\text{TIP}_{22}$  & 46.47 & 79.74  &50.09  &61.44  &89.86  & 15.36 &  53.96 & 80.61 & 61.15 & 50.62 & 76.54 & 20.60 & 45.18 & 76.32 & 24.87 & 69.88 & 93.43 & 20.22   \\
SCTransNet~\cite{SCTransNet} & $\text{TGRS}_{24}$ & 49.74 & 83.44 &  53.24& 62.80 & 90.99 & 16.79&  51.55 & 80.08 & 81.75 & 44.38 & 71.65 & 32.54 & 41.83 & 73.48 & 70.91 & 70.57 & 95.87 & 39.48   \\
MSHNet~\cite{liu2024infrared}  & $\text{CVPR}_{24}$ & 40.36 & 83.10 &  123.38& 61.78 & 91.93 & 22.69&  52.03 & 80.48 & 67.43 & 49.36 & 76.06 & 35.22 & 42.39 & 71.28 & 38.88 & 66.42 & 93.80 & 15.36  \\
P-MSHNet~\cite{Pinwheel}  & $\text{AAAI}_{25}$ & 44.41 & 80.95 &  92.81& 62.30 & 92.49 & 19.94 & 52.55 & 80.34  & 71.15 & 46.96 & 76.59 & 63.13  & 37.31 & 66.12 & 55.92 & 70.09 & 94.55 & 12.39   \\
BGM~\cite{liu2025forgetting} & $\text{TGRS}_{25}$ &  50.82 & 82.43 &  83.16& 61.95 & 92.49 & 14.29 & 44.44 & 75.57 & 112.34 & 41.56 & 72.24 & 50.53 & 39.71 & 72.57 & 40.09 & 67.74 & 93.62 & 12.75    \\
PAL~\cite{yu2025easyhard} & $\text{ICCV}_{25}$ & 46.90 & 81.22 &  58.94& 59.88 & 91.36 & 10.86 &  50.83  & 79.27 & 55.56 & 44.03 & 74.55 & 34.88 & 41.03  & 75.52 & 64.32 & 67.60  & 95.87 & 31.73   \\
SGIRNet~\cite{chi2025contrast} & $\text{KBS}_{25}$ &   28.34 & 76.11 &  60.52& 47.40 & 84.17 & 67.64 &  ---  & ---  & ---  & ---  & ---  & ---  & ---  & ---  & ---  & ---  & ---  & ---   \\
HSDWNet~\cite{Liu2026HSDW} & $\text{SPL}_{26}$ &   41.23 & 79.59 &  86.92 & 58.87 & 92.66 & 21.11 &  45.04  & 78.23  & 72.50  & 43.18  & 73.12  & 41.80  & 42.14  & 74.49 & 58.67  & 64.03  & 95.41  & 35.31   \\
\midrule\rowcolor{cyan!6}
\textbf{S$^2$CP-SCTransNet} & --- &   \textbf{53.76} & \textbf{87.88} & \textbf{49.25} & \textbf{66.25} & 95.30 & \textbf{10.82} & 54.50 & \textbf{81.49} & 56.08 & \textbf{51.98} & \textbf{77.61}  & 30.55 & \textbf{48.08} & \textbf{76.97} & 28.74 & 72.71 & 95.12  & \textbf{12.00}  \\
\rowcolor{cyan!6}
\textbf{S$^2$CP-MSHNet} & --- & 51.50 & 87.21 & 52.97 & 64.96 & \textbf{95.49} & 12.61 & \textbf{54.70}  & 80.14 & \textbf{39.51} & 51.07 & 76.81 & \textbf{16.43} & 46.81  & 75.20 & \textbf{23.10} & \textbf{72.78} & \textbf{96.06} & 13.54  \\
\bottomrule
\end{tabular} 
\label{tab:sdg}
\end{table*}

\subsubsection{Implementation Details}
In our experiments, a 4-stage encoder–decoder architecture is adopted as the backbone network, which is trained for 300 epochs using the Adam optimizer with the Soft-IoU loss~\cite{huang2019batching}.
The initial learning rate and batch size are set to $5\mathrm{e}{-4}$ and 8, respectively.
The hyperparameters $\tau$ and $\lambda$ involved in selective style recomposition are all set to 0.3.
The images in all datasets are resized to $256 \times 256$.
All implementations are conducted based on the PyTorch framework with NVIDIA RTX 3090 GPU support.

\subsection{Experimental Results}
We compare the cross-domain detection performance of S$^2$CPNet against various state-of-the-art IRSTD methods, including six model-driven methods (Tophat~\cite{tophat}, MPCM~\cite{MPCM}, RLCM~\cite{RLCM}, LIG~\cite{LIG}, TLLCM~\cite{TLLCM}, and MSLSTIPT~\cite{MSLSTIPT}) and eighteen data-driven methods (ACMNet~\cite{ACMNet}, DNANet~\cite{DNANet}, UIUNet~\cite{UIUNet}, ISNet~\cite{ISNet}, AGPCNet~\cite{AGPCNet}, ABC~\cite{ABC}, MTUNet~\cite{MTUNet}, SCTransNet~\cite{SCTransNet}, SeRankDet~\cite{SeRankDet}, HCFNet~\cite{HCFNet}, L2SKNet~\cite{L2SKNet}, MSHNet~\cite{liu2024infrared}, DATransNet~\cite{DAtransNet}, P-MSHNet~\cite{Pinwheel}, BGM~\cite{liu2025forgetting}, PAL~\cite{yu2025easyhard}, SGIRNet~\cite{chi2025contrast}, and HSDWNet~\cite{Liu2026HSDW} ).
Since many IRSTD architectures and learning strategies can be flexibly combined, following common practice in prior works, we implement two variants of our spatial–spectral collaborative perception paradigm: S$^2$CP-SCTransNet and S$^2$CP-MSHNet.
\subsubsection{Quantitative Analysis}
We comprehensively evaluate the cross-domain IRSTD performance of different methods on both single-domain and multi-domain generalization scenarios. The performance is reported by metrics at both the pixel level (intersection over union (IoU) and F-measure ($F_{1}$)) and the target level (probability of detection ($P_d$), false alarm rate ($F_a$), and receiver operating characteristic (ROC) curve).

\begin{table*}[t]
\footnotesize 
\renewcommand\arraystretch{1.1}
\setlength{\tabcolsep}{6.5pt}
\centering
\caption{Comparison of cross-domain ISTD performance in the multi-domain generalization scenario on NUAA-SIRST, NUDT-SIRST, and IRSTD-1K datasets in terms of IoU (\%), $F_{1}$ (\%), $P_{d}$ (\%), and $F_{a}$ ($10^{-6}$). The best results are highlighted in bold.}
\begin{tabular}{l|c| c c c c| c c c c| c c c c}
\toprule
\multirow{2}{*}{Method} & \multirow{2}{*}{Venue} &
\multicolumn{4}{c|}{NUAA-SIRST} &
\multicolumn{4}{c|}{NUDT-SIRST} &
\multicolumn{4}{c}{IRSTD-1K} \\
\cline{3-14}
& & IoU $\uparrow$ &  $F_1 \uparrow$  &  $P_d\uparrow$ &   $F_a\downarrow$
& IoU $\uparrow$ &  $F_1 \uparrow$  &  $P_d\uparrow$ &   $F_a\downarrow$
& IoU $\uparrow$ &  $F_1 \uparrow$  &  $P_d\uparrow$ &   $F_a\downarrow$ \\
\midrule
Tophat~\cite{tophat}       & $\text{OE}_{96}$ & 27.80 & 43.51 & 90.74 & 193.67 & 12.59 & 22.37 & 69.41 & 76.50  &  5.95 & 11.23 & 69.69 & 621.38 \\
MPCM~\cite{MPCM}           & $\text{PR}_{16}$  & 45.18 & 62.24 & 90.74 & 26.25  & 15.60 & 26.98 & 58.62 & 185.54 & 22.90 & 37.27 & 79.12 & 135.53 \\
RLCM~\cite{RLCM}           & $\text{GRSL}_{18}$ & 27.16 & 42.71 & 84.25 & 135.79 & 15.38 & 26.66 & 76.40 & 480.03 & 17.67 & 30.04 & 64.64 & 54.05 \\
LIG~\cite{LIG}             & $\text{IPT}_{18}$& 30.10 & 46.27 & 88.88 & 77.39  & 28.29 & 36.20 & 78.41 & 170.65 & 11.67 & 20.90 & 81.14 & 624.59 \\
TLLCM~\cite{TLLCM}         & $\text{GRSL}_{19}$ & 17.92 & 30.40 & 84.25 & 20.04  & 11.77 & 21.07 & 72.59 & 151.83 & 11.28 & 27.67 & 81.81 & 212.67 \\
MSLSTIPT~\cite{MSLSTIPT}   & $\text{TGRS}_{20}$ & 19.80 & 33.05 & 83.33 & 62.71  &  9.47 & 17.29 & 59.89 & 104.26 & 17.82 & 30.25 & 61.61 & 91.32 \\\midrule 
ACMNet~\cite{ACMNet}       &$\text{WACV}_{21}$ & 64.39 & 78.34 & 95.41 & 26.25  & 37.85 & 54.91 & 71.95 & 56.07  & 42.64 & 59.78 & 83.33 & 117.28 \\
DNANet~\cite{DNANet}       & $\text{TIP}_{22}$ & 66.59 & 79.94 & 95.42 & 17.56  & 49.08 & 65.84 & 79.15 & 42.88  & 48.81 & 65.59 & 86.73 & 106.12 \\
UIUNet~\cite{UIUNet}       & $\text{TIP}_{22}$  & 64.27 & 78.25 & 92.66 & 25.19   & 47.86 & 64.73 & 78.20 & 48.35  & 44.81 & 61.88 & 87.41 & 151.14 \\
ISNet~\cite{ISNet}       & $\text{CVPR}_{22}$  &65.45  &79.12  & 94.49 & 14.01 & 44.53 &61.73  &75.34  & 49.91  & 39.62 &56.75  & 85.71 &  156.07  \\
AGPCNet~\cite{AGPCNet}     & $\text{TAES}_{23}$ & 65.93 & 79.47 & 95.41 & 10.46  & 42.76 & 59.91 & 75.02 & 39.89  & 48.33 & 65.16 & 84.01 & 94.13 \\
ABC~\cite{ABC}             & $\text{ICME}_{23}$ & 68.98 & 81.64 & 96.33 & 8.51   & 47.48 & 64.39 & 77.88 & 39.34  & 43.28 & 60.42 & 87.07 & 163.06 \\
MTUNet~\cite{MTUNet}       & $\text{TGRS}_{23}$ & 66.53 & 79.90 & 93.57 & 15.96  & 43.62 & 60.74 & 75.02 & 68.02  & 47.63 & 64.52 & 87.42 & 116.30 \\
SCTransNet~\cite{SCTransNet}& $\text{TGRS}_{24}$ & 66.84 & 80.12 & 95.41 & 19.51  & 44.95 & 63.26 & 76.71 & 63.26  & 50.16 & 66.80 & 82.31 & 85.68 \\
SeRankDet~\cite{SeRankDet} & $\text{TGRS}_{24}$ & 67.87 & 80.86 & 97.24 & 12.06  & 52.33 & 68.50 & 81.69 & 56.53  & 44.48 & 61.57 & 87.75 & 123.05 \\
HCFNet~\cite{HCFNet}       & $\text{ICME}_{24}$ & 69.58 & 82.06 & 97.24 & 14.19   & 48.65 & 65.45 & 81.69 & 70.82  & 52.86 & 69.16 & 86.39 & 75.30 \\
L2SKNet~\cite{L2SKNet}     & $\text{TGRS}_{24}$ & 68.90  & 81.58 & 96.33 & 7.98 & 47.54 & 64.44 & 80.63 & 106.42 & 52.73  & 69.04 & 86.39 & 73.78\\
MSHNet~\cite{liu2024infrared}   & $\text{CVPR}_{24}$ & 67.29 & 80.45 & 95.41 & 15.25 & 48.02 & 64.88 & 78.41 & 38.44 & 49.42 & 66.14 & 86.73  & 111.67  \\
DATransNet~\cite{DAtransNet}& $\text{GRSL}_{25}$ & 67.52 & 80.61 & 96.33 & 8.69   & 46.57 & 65.45 & 77.46 & 39.47  & 47.21 & 64.13 & 86.05 & 80.08 \\
P-MSHNet~\cite{Pinwheel}   & $\text{AAAI}_{25}$ & 68.11 & 81.03 & 96.33 & 6.74   & 46.60 & 63.57 & 78.30 & 44.19  & 49.40 & 66.12 & 85.71 & 87.07 \\
BGM~\cite{liu2025forgetting} & $\text{TGRS}_{25}$ & 67.09 & 80.30 & 95.41 & 8.33 & 45.74 & 62.76 & 77.24 & 40.49 & 48.08 & 64.92 & 79.25 & 80.77  \\
PAL~\cite{yu2025easyhard}& $\text{ICCV}_{25}$ & 66.74 & 80.05 & 96.33 & 22.71 & 44.40 & 61.50 & 77.98 & 44.76 & 48.54 & 65.33 & 80.27 & 85.47 \\
\midrule \rowcolor{cyan!6}
\textbf{S$^2$CP-SCTransNet} & --- & \textbf{71.73} & \textbf{83.53} & \textbf{99.08} & 8.33   & \textbf{56.85} & \textbf{72.49} & \textbf{85.39} & \textbf{28.88}  & 59.30 & 74.34 & 87.07 & \textbf{37.27} \\ \rowcolor{cyan!6}
\textbf{S$^2$CP-MSHNet} & --- &70.47  & 82.67 & 97.24 &\textbf{2.30}   & 55.71 & 71.55 & 82.53 & 32.33  & \textbf{60.01} & \textbf{74.99} & \textbf{90.47} & 41.90 \\
\bottomrule
\end{tabular} 
\label{tab:com_result}
\end{table*}

\begin{figure*}[t]
\centering  
\subfloat[NUAA-SIRST]{\includegraphics[width=0.33\textwidth]{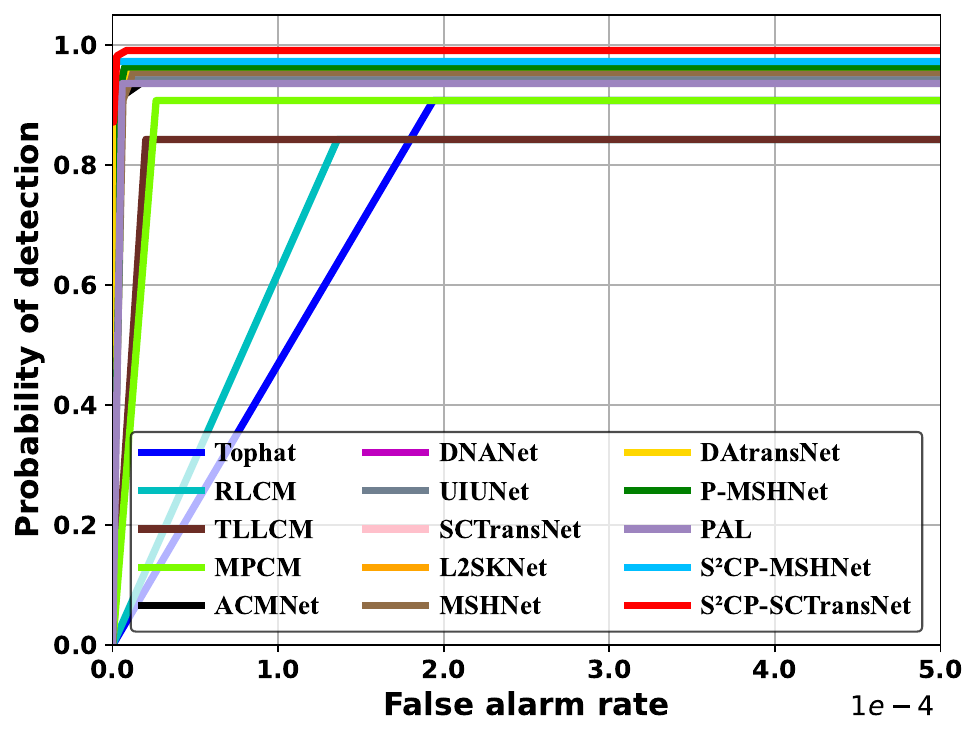}} 
\subfloat[NUDT-SIRST]{\includegraphics[width=0.33\textwidth]{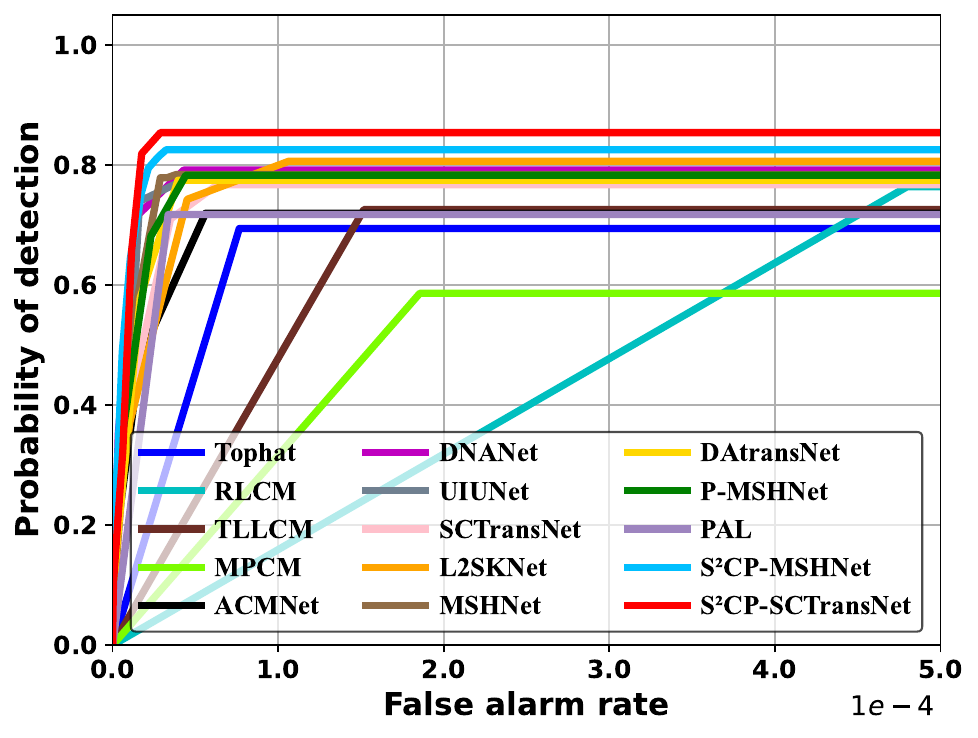}} 
\subfloat[IRSTD-1K]{\includegraphics[width=0.33\textwidth]{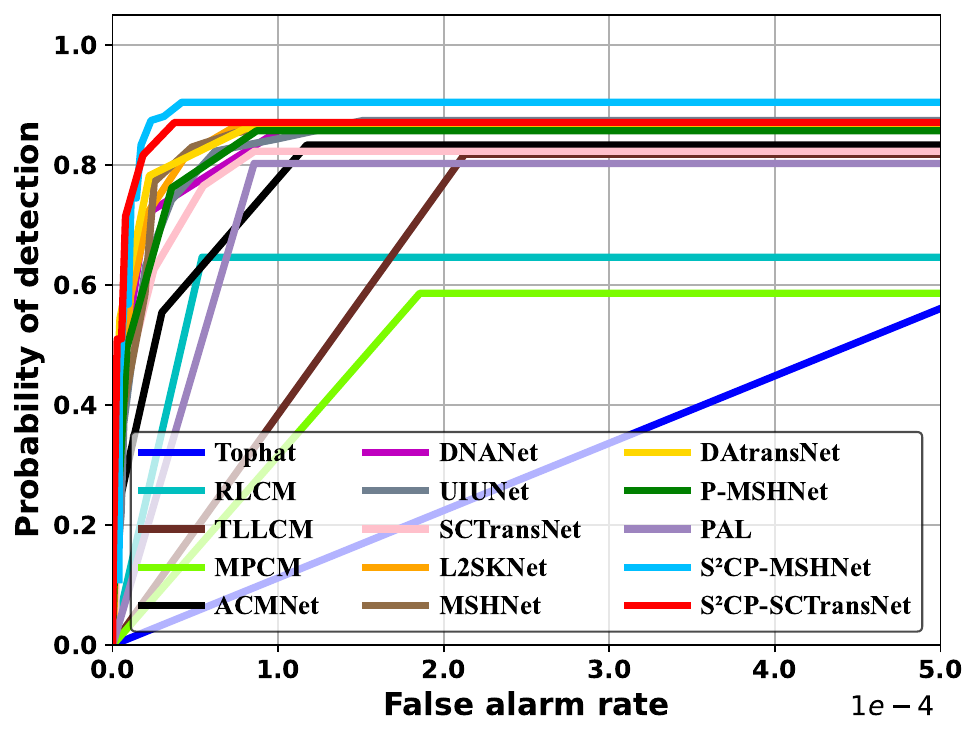}}
\caption{ROC curves of different methods when generalized to NUAA-SIRST, NUDT-SIRST, and IRSTD-1K datasets.}
\label{istd_roc}
\end{figure*}

The quantitative results of different methods on cross-domain IRSTD are reported in Tables~\ref{tab:sdg}~and~\ref{tab:com_result}.
According to the results, our method demonstrates clear superiority over comparison methods across different scenarios and domains.
First, in the single-domain generalization scenario without applying SSR, the proposed method achieves more accurate cross-domain detection performance at both pixel and object levels.
Notably, when integrated with the same SCTransNet framework, S$^2$CPNet surpasses the latest specialized single-domain generalized method, SGIRNet, by a substantial margin.
Then, in the multi-domain generalization scenario, the proposed method improves the overall performance of the integrated baseline frameworks (SCTransNet and MSHNet) by approximately 8\% in IoU and 6\% in F-measure. 
Meanwhile, the target perceptibility is effectively enhanced while preventing more than half of the false alarms.
In addition, applying learning strategies designed for single-domain settings not only fails to provide performance gains but can even cause degradation, further underscoring the necessity of rethinking representations for cross-domain IRSTD.
Moreover, to thoroughly assess detection performance under varying thresholds, we also plot the ROC curves of different methods in Fig.~\ref{istd_roc}. 
It can be observed that the curves of our method rise more rapidly and enclose larger areas than those of other methods, indicating a better trade-off between detection accuracy and false-alarm control.

\begin{figure*}[]
\centering
\includegraphics[width=\textwidth]{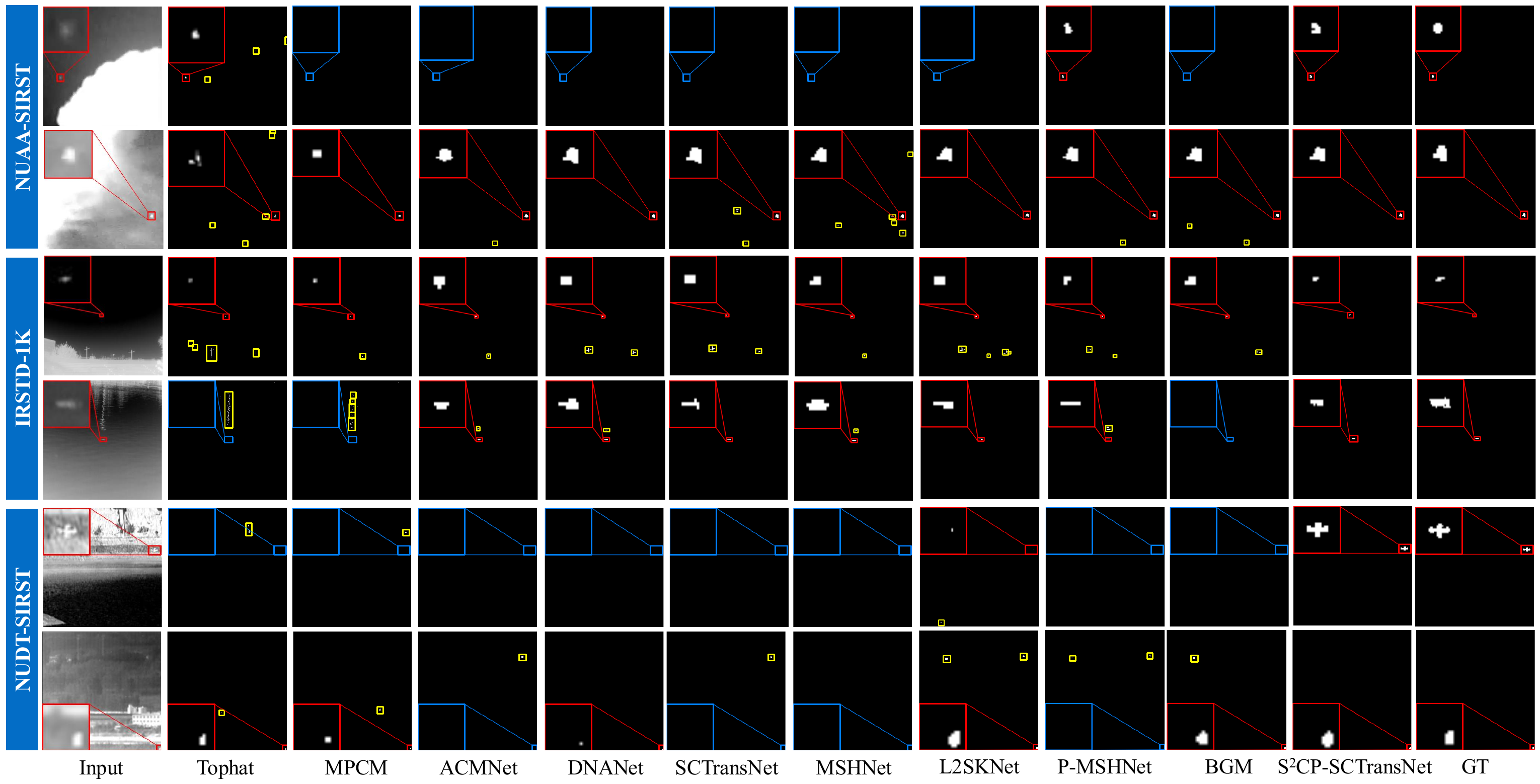}
\caption{Visualizations of cross-domain IRSTD results of different methods. Boxes in red, blue, and yellow refer to detected targets, miss detections, and false alarms. Zoomed-in regions of the detected targets are provided in the corner for clear visualization.}
\label{vis_mask}
\end{figure*}

\subsubsection{Qualitative Analysis}
Beyond the quantitative analysis above, we also conduct qualitative analysis to provide a more intuitive illustration of the performance comparison of our method and existing IRSTD methods.
The visualization results across different target domains under the multi-domain generalization setting are shown in Fig.~\ref{vis_mask}.
We first observe that, although model-driven methods are insensitive to distribution shifts, their limited representation ability still results in numerous false alarms and missed detections.
Compared with other data-driven IRSTD methods, S$^2$CPNet can more accurately distinguish true targets within the sensing field while effectively avoiding false activations caused by background clutter and noise.
In addition, even when a target is heavily entangled with background clutter in unseen domains (as shown in the second-to-last row), the proposed method can still accurately identify it from complex backgrounds, whereas it is almost completely overlooked by other methods.
Moreover, for the correctly detected targets across different domains, the proposed method can more precisely delineate their morphological details.

\begin{table}[]
\footnotesize
\renewcommand\arraystretch{1}
\setlength{\tabcolsep}{6pt}
\centering
\caption{Ablation experiments about contributions of each component to cross-domain IRSTD. Best results are highlighted in bold.}
\begin{tabular}{ c c c| c c c c}
\toprule
PRM& OAM & SSR & IoU $\uparrow$ &  $F_1 \uparrow$ & $P_d\uparrow$ &   $F_a\downarrow$ \\
\midrule
\ding{56}& \ding{56} &\ding{56} &  49.42 & 66.14 & 86.73  & 111.67  \\
\ding{52}& \ding{56} &\ding{56}&  57.64 & 73.12 & 87.75 & 38.64  \\
\ding{56}& \ding{52} &\ding{56}&  51.93 & 68.35 & 89.45 & 92.01  \\
\ding{56}& \ding{56} &\ding{52} &  54.47 & 70.05 & 88.77 & 66.50  \\
\ding{56}& \ding{52} &\ding{52} &  55.62 & 71.47 & 85.71 & 63.69  \\
\midrule \rowcolor{cyan!6}
\ding{52}& \ding{52} & \ding{52}&  \textbf{60.01} & \textbf{74.99} & \textbf{90.47} & \textbf{41.90}  \\
\bottomrule
\end{tabular}
\label{tab:ab}
\end{table}

\subsection{Ablation Studies}
\subsubsection{Contribution of Each Component}
The proposed S$^2$CPNet consists of three key components: PRM, OAM, and SSR. 
To evaluate the contributions of each component to cross-domain IRSTD, we conduct ablation experiments under the multi-domain generalization setting, where the IRSTD-1K dataset is served as the target domain. 
Specifically, we adopt MSHNet as the baseline and evaluate the impact of different component combinations on cross-domain performance in Table~\ref{tab:ab}.
First, we find that integrating PRM with convolutional blocks alone can lead to a notable performance improvement, which validates our frequency-domain insight into domain discrepancies.
Moreover, the individual use of OAM and SSR also provides certain performance gains. However, the improvement brought by OAM remains limited due to the lack of generalized encoded representations.
Even when OAM and SSR are combined, a clear performance gap remains compared with the full S$^2$CPNet, which further confirms the effectiveness of spatial–spectral collaborative perception in cross-domain IRSTD.

\begin{figure}[]
\centering  
\subfloat[IoU]{\includegraphics[width=0.25\textwidth]{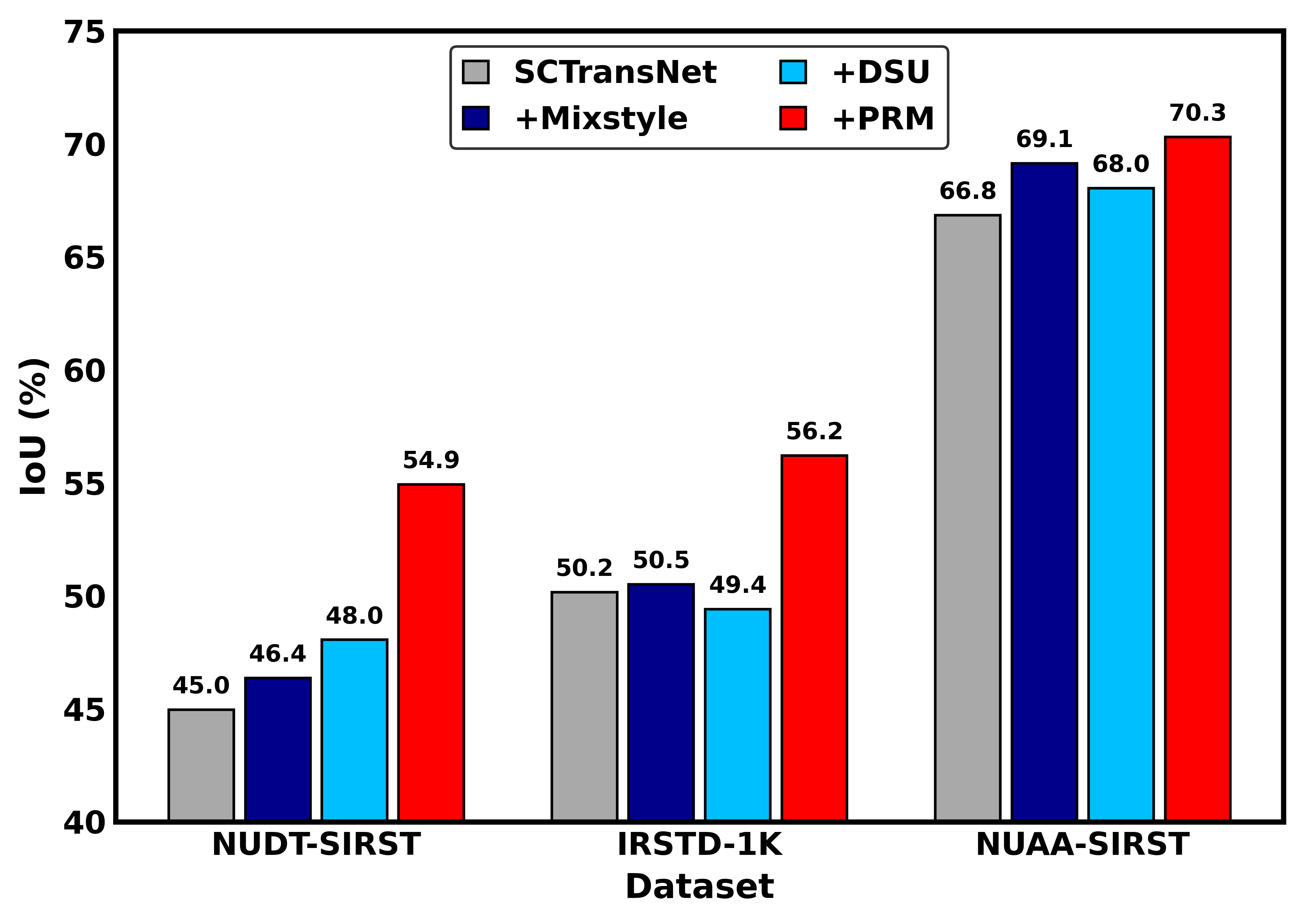}} 
\subfloat[$P_d$]{\includegraphics[width=0.25\textwidth]{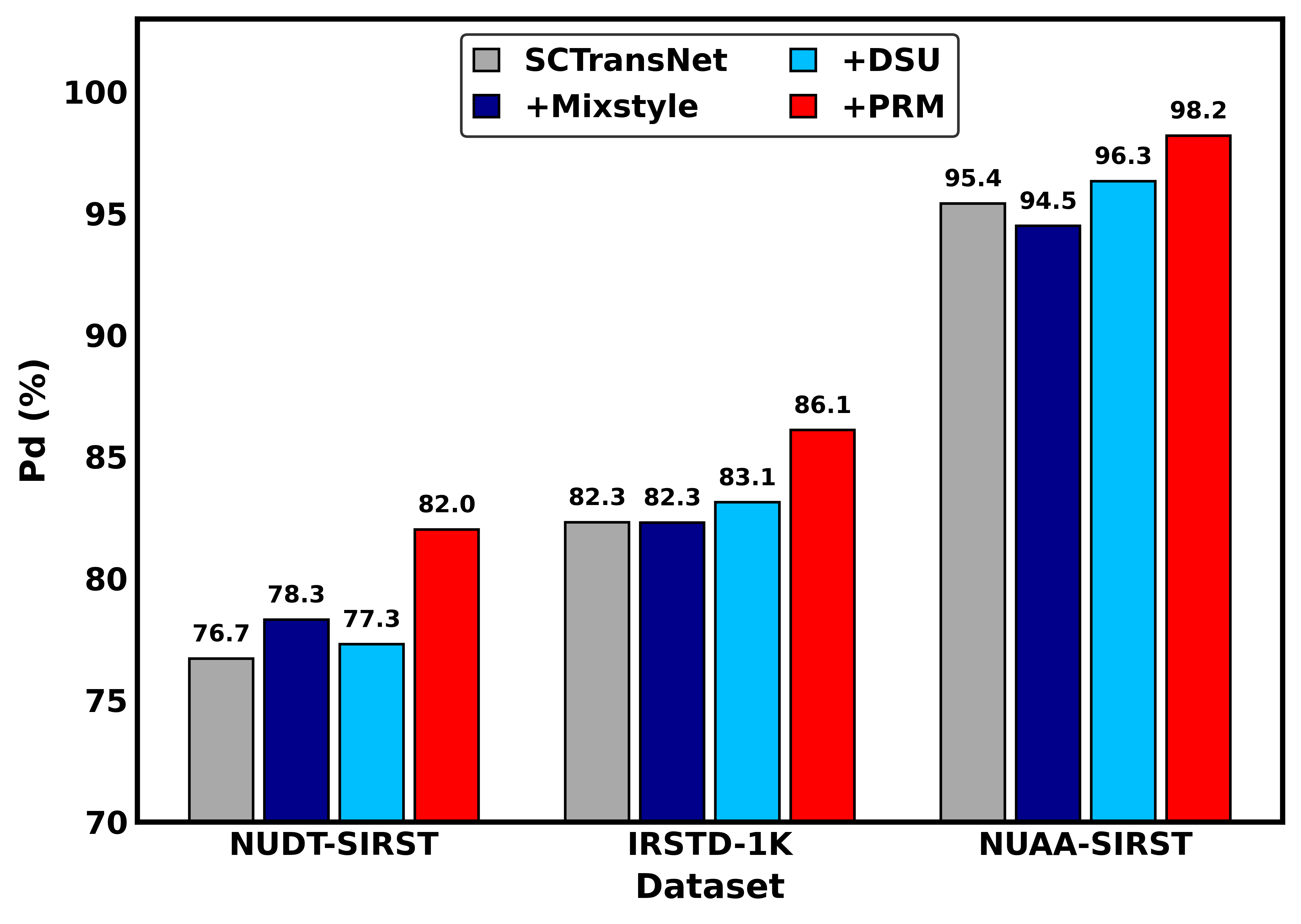}} 
\caption{Cross-domain IRSTD performance of SCTransNet combined with different DG approaches.}
\label{ab_freq}
\end{figure}

\subsubsection{Frequency Domain Rectification}
As the core of our proposed method, we mitigate domain discrepancies from a frequency perspective to derive generalizable target awareness.
To verify the rationality of this design, we compare the efficacy of PRM with two commonly adopted DG approaches, MixStyle~\cite{zhoudomain} and domain shifts with uncertainty (DSU)~\cite{liuncertainty}, in cross-domain IRSTD.
The performance is reported by IoU and $P_d$ in Fig.~\ref{ab_freq}, where SCTransNet is adopted as the baseline model.
According to the results, directly applying DG approaches designed for natural images within conventional spatial learning pipelines brings only limited performance improvement.
In contrast, unifying the phase congruency of representations in the frequency domain effectively boosts IRSTD performance across domains, and the superiority becomes more evident on challenging target domains (NUDT-SIRST and IRSTD-1K).

\subsubsection{Orthogonal Attention Mechanism}
\begin{figure*}[]
\centering
\includegraphics[width=\textwidth]{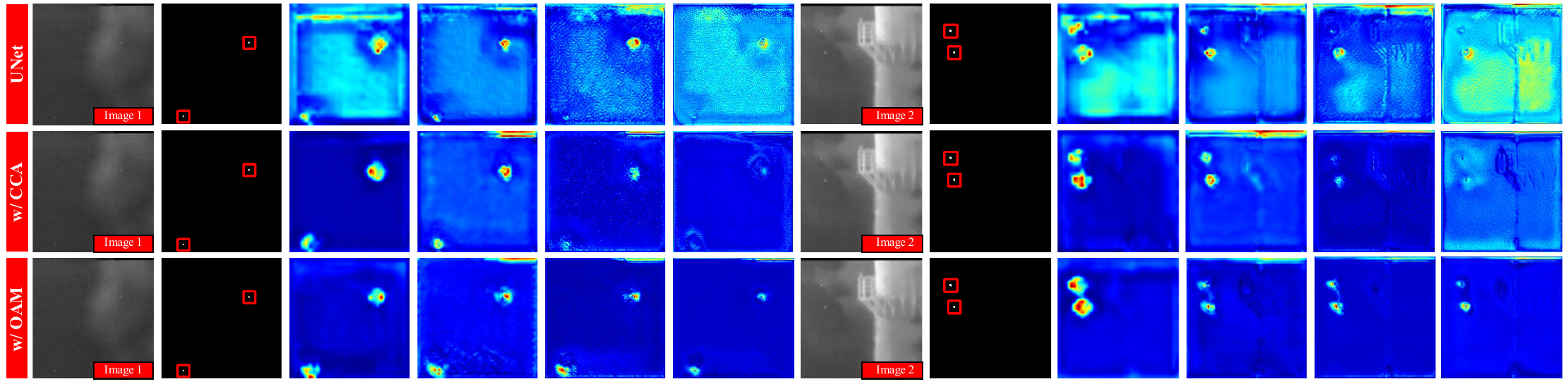}
\caption{Visualization of heatmaps during different decoding stages when applying CCA and OAM to UNet.}
\label{heatmap_ab}
\end{figure*}

During decoding, we employ two orthogonal importance descriptors to address the positional information loss introduced by existing attention mechanisms when refining cross-stage representations.
To verify the effectiveness of this design, we compare heatmaps at different decoding stages when applying channel-wise cross-attention (CCA)~\cite{wang2022uctransnet} and OAM on the same U-Net backbone.
The visualization results are provided in Fig.~\ref{heatmap_ab}.
As shown in the first row, when using only the original skip connections, the semantic misalignment between encoder and decoder features results in disorganized attention to irrelevant regions.
Although applying CCA helps suppress background interference in the early decoding stages, the gradual positional information loss introduced by the 2-D global pooling operation ultimately results in the disappearance of target responses.
In comparison, inserting OAM into skip connections can effectively refine cross-stage representations while preserving perceptual sensitivity to target localization.

\begin{figure}[]
\centering 
\subfloat[IoU]{\includegraphics[width=0.25\textwidth]{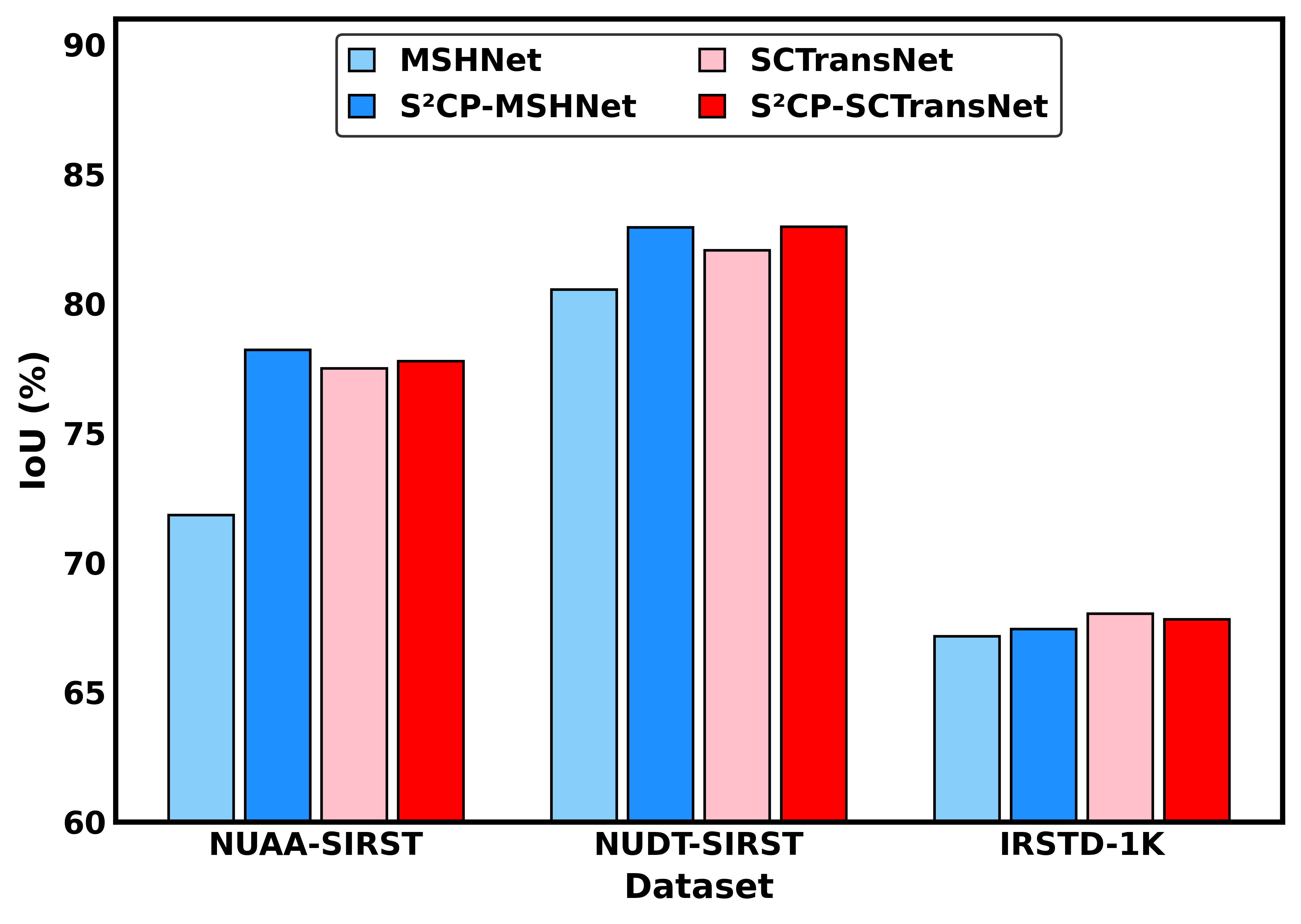}} 
\subfloat[$P_d$]{\includegraphics[width=0.25\textwidth]{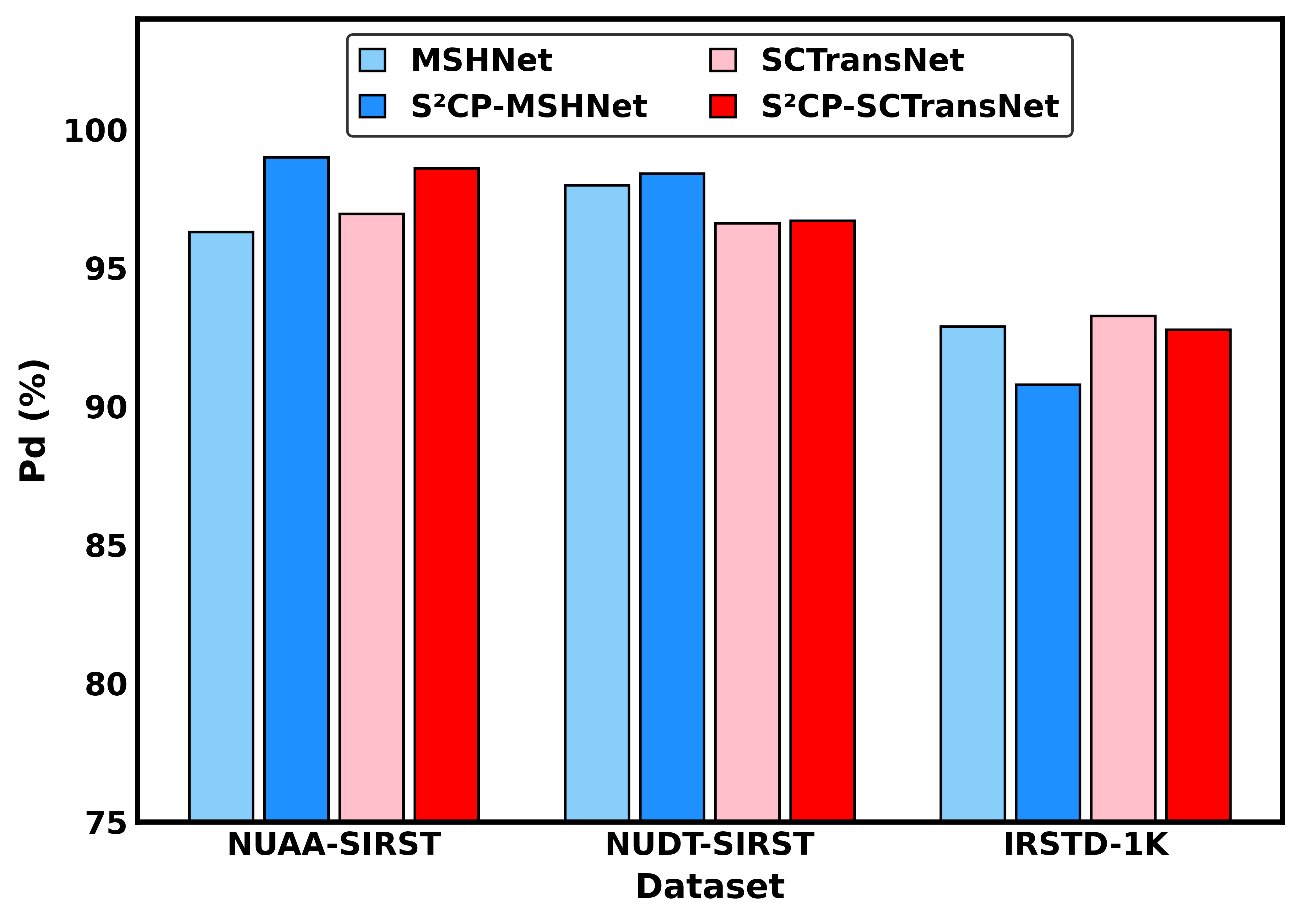}} 
\caption{IRSTD performance on different datasets under domain-consistent setting.}
\label{fig:conven} 
\end{figure}

\subsubsection{Performance under Conventional Setup}
In this experiment, we further assess the effectiveness of the proposed S$^2$CPNet under the conventional IRSTD setup.
Specifically, we adopt the domain-consistent setting in which each model is trained and evaluated on the corresponding training and test splits of the same dataset, i.e., under identical data distributions.
We compare the two variants of S$^2$CPNet with their respective state-of-the-art baselines: SCTransNet and MSHNet, across all three datasets.
The performance is reported by IoU and $P_d$ in Fig.~\ref{fig:conven}, we observe that S$^2$CPNet generally preserves the performance of the baseline methods under the standard training–testing setup, and even brings additional improvements in some cases.
These improvements are mainly attributable to the incorporation of PRM during encoding, which facilitates the derivation of generalizable target awareness, and the integration of OAM into skip connections, which preserves critical positional information throughout decoding.

\subsubsection{Computational Efficiency}
We analyze the computational efficiency of S$^2$CPNet by comparing the computational complexity and overall detection performance of its variants and the respective baseline methods.
We assess computational efficiency using two widely used metrics in IRSTD: the model parameter count and the floating-point operations (FLOPs).
As reported in Table~\ref{ab_com}, the proposed method achieves substantial improvements in cross-domain IRSTD performance without incurring excessive computational overhead, which demonstrates its high computational efficiency.

\begin{table}[] 
\caption{Analysis of computational efficiency.}
\label{ab_com} 
\footnotesize
\renewcommand\arraystretch{1}
\setlength{\tabcolsep}{5.5pt}
\centering
\begin{tabular}{l|c|c|ccc}
\toprule
Methods & Params & FLOPs  &IoU $\uparrow$& $P_d\uparrow$ & $F_a\downarrow$  \\ 
\midrule
MSHNet & 4.1M&  6.1G &  54.91 &86.85 &55.12 \\
\rowcolor{cyan!6}
S$^2$CP-MSHNet& 5.5M&  6.7G &62.06   & 90.08&25.51 \\
SCTransNet & 11.2M&  10.1G & 53.98  & 84.81&56.15 \\  \rowcolor{cyan!6}
S$^2$CP-SCTransNet& 12.3M&  11.3G &62.62   & 90.51& 24.82 \\
\bottomrule
\end{tabular} 
\end{table}

\begin{table}[] 
\caption{Detection performance with different parameter settings of SSR. Best results are highlighted in bold.}
\label{hyper} 
\footnotesize
\renewcommand\arraystretch{0.95}
\setlength{\tabcolsep}{6pt}
\centering
\begin{tabular}{c|c|cccc}
\toprule 
Parameter & Setting & IoU $\uparrow$ &  $F_1 \uparrow$ & $P_d\uparrow$ &   $F_a\downarrow$ \\ \midrule
\multirow{4}{*}{$\tau$} &0.2 &  54.46 &  70.51& 81.58 & 37.45\\
&\cellcolor{cyan!6}0.3  & \cellcolor{cyan!6}\textbf{55.35}& \cellcolor{cyan!6}\textbf{71.26} & \cellcolor{cyan!6}\textbf{82.75} & \cellcolor{cyan!6}30.21\\
&0.4  &  54.50 &  70.55 & 82.10 & \textbf{28.33} \\
&0.5  &  53.17 &  69.43 & 80.52 & 38.28 \\
\midrule 
\multirow{5}{*}{$\lambda$} &0.1 &55.64   & 71.49 &
82.53  &29.75 \\
&0.2  & 56.25  & 72.01 & 84.76 & 29.82\\
&\cellcolor{cyan!6}0.3  & \cellcolor{cyan!6}\textbf{56.85}  &\cellcolor{cyan!6}\textbf{72.49} & \cellcolor{cyan!6}\textbf{85.39} & \cellcolor{cyan!6}\textbf{28.88}\\
&0.4  & 56.83 &  72.47& 84.65 & 31.06\\
&0.5  & 55.97  &  71.77& 82.85 &36.53 \\
\bottomrule
\end{tabular} \label{tab:para} 
\end{table}

\subsubsection{Parameters in Selective Style Recomposition}
We investigate the impact of different parameter settings in SSR on cross-domain IRSTD performance, including $\tau$ and $\lambda$.
Following common practice in plug-and-play DG modules~\cite{zhoudomain,huang2023style}, we apply SSR only to the first two residual blocks, where domain-related information is mainly encoded.
First, we fix $\lambda$ to 0 and evaluate the cross-domain performance of S$^2$CP-SCTransNet on the NUDT-SIRST dataset with varying $\tau$.
As reported in Table~\ref{tab:para}, a small $\tau$ results in weak regularization of domain gaps, whereas an excessively large one tends to degrade performance, and a moderate value achieves the best trade-off.
Then, we fix $\tau$ at 0.3 and vary $\lambda$ from 0.1 to 0.5.
The results indicate that preserving part of the original representations benefits detection performance, and the model remains relatively insensitive to changes in $\lambda$ within a certain range unless it is excessively large.
Therefore, we set both $\tau$ and $\lambda$ to 0.3.

\section{Conclusion}\label{sec5}
In this paper, we propose S$^2$CPNet, a spatial–spectral collaborative perception framework for cross-domain IRSTD.
We rethink representations from a frequency perspective, providing an insightful revelation of domain discrepancies in IRSTD, upon which a phase rectification module is incorporated into the encoding stage to improve generalization.
During decoding, the deployment of the orthogonal attention mechanism effectively mitigates the positional degradation when refining cross-stage representations.
Moreover, overfitting to domain-specific patterns is further alleviated through selective style recomposition.
Extensive experiments across diverse cross-domain IRSTD scenarios fully validate the superiority of our proposed method, and the established benchmarks offer a valuable groundwork for future research.

\bibliographystyle{IEEEtran}
\bibliography{Reference}

\end{document}